\documentclass[twocolumn,10pt]{article}

\usepackage[utf8]{inputenc}
\usepackage[T1]{fontenc}
\usepackage[a4paper,top=1.9cm,bottom=2.0cm,left=1.7cm,right=1.7cm,
            columnsep=0.7cm]{geometry}
\usepackage{lmodern}                  
\usepackage{mathptmx}                 
\usepackage{microtype}
\usepackage{graphicx}
\usepackage[table,dvipsnames]{xcolor} 
\usepackage{booktabs}
\usepackage{tabularx}
\usepackage{pdflscape}
\usepackage{ragged2e}
\usepackage{caption}
\usepackage{fancyhdr}
\usepackage[super,comma,sort&compress]{natbib} 
\usepackage{hyperref}
\hypersetup{colorlinks=true, linkcolor=blue!45!black,
            citecolor=blue!45!black, urlcolor=blue!45!black}

\definecolor{bioCol}{HTML}{2A9D8F}
\definecolor{humanCol}{HTML}{3A6EA5}
\definecolor{machineCol}{HTML}{C1495B}
\definecolor{accentCol}{HTML}{3A6EA5}
\definecolor{preserveCol}{HTML}{2F9E6F}
\definecolor{substituteCol}{HTML}{C1495B}
\definecolor{heatInst}{HTML}{BFE6CF}
\definecolor{heatDisp}{HTML}{F6DCA0}
\definecolor{heatUnt}{HTML}{E6E6EC}
\definecolor{heatUntInk}{HTML}{9A9AA8}
\definecolor{headerDark}{HTML}{33333F}
\definecolor{zebraCol}{HTML}{FAFAFC}
\definecolor{rowHeadCol}{HTML}{F4F4F8}
\definecolor{preTint}{HTML}{EEF8F1}
\definecolor{subTint}{HTML}{FBEEEE}

\newcommand{\hd}[2]{\cellcolor{#1}\textcolor{white}{\sffamily\bfseries #2}}
\newcommand{\rh}[1]{\cellcolor{rowHeadCol}\textbf{#1}}
\newcommand{\gInst}{\cellcolor{heatInst}\textbf{\textbullet}}
\newcommand{\gDisp}{\cellcolor{heatDisp}\textbf{\textasciitilde}}
\newcommand{\gUnt}{\cellcolor{heatUnt}\textcolor{heatUntInk}{\textperiodcentered}}
\newcommand{\gInstC}{\cellcolor{heatInst}\textbf{\textbullet}\rlap{\,\textsuperscript{\dag}}}
\newcommand{\gDispC}{\cellcolor{heatDisp}\textbf{\textasciitilde}\rlap{\,\textsuperscript{\dag}}}

\DeclareCaptionLabelSeparator{bar}{ {\textbar}\ }
\makeatletter
\renewcommand\section{\@startsection{section}{1}{\z@}%
  {-2.2ex \@plus -0.6ex \@minus -.2ex}{0.9ex \@plus .2ex}%
  {\sffamily\large\bfseries\color{headerDark}}}
\renewcommand\subsection{\@startsection{subsection}{2}{\z@}%
  {-1.6ex \@plus -0.5ex}{0.5ex \@plus .2ex}%
  {\sffamily\normalsize\bfseries\color{headerDark}}}
\makeatother

\newcounter{lead}[section]
\renewcommand{\thelead}{\thesection.\arabic{lead}}

\newcommand{\lead}[1]{%
  \refstepcounter{lead}%
  \par\medskip\noindent
  {\sffamily\bfseries\thelead\ #1.}%
  \hspace{0.6em}%
}

\definecolor{diagCol}{HTML}{6B5B95}
\newcolumntype{D}{>{\hsize=0.9\hsize\RaggedRight\arraybackslash}X}
\newcolumntype{F}{>{\hsize=0.95\hsize\RaggedRight\arraybackslash}X}
\newcolumntype{E}{>{\hsize=0.85\hsize\RaggedRight\arraybackslash}X}
\newcolumntype{M}{>{\hsize=1.2\hsize\RaggedRight\arraybackslash}X}
\newcolumntype{G}{>{\hsize=1.0\hsize\RaggedRight\arraybackslash}X}
\newcolumntype{R}{>{\hsize=1.15\hsize\RaggedRight\arraybackslash}X}
\newcolumntype{K}{>{\hsize=1.05\hsize\RaggedRight\arraybackslash}X}

\fancypagestyle{titlestyle}{\fancyhf{}%
  \fancyfoot[C]{\footnotesize\sffamily\thepage}}

\begin{document}

\twocolumn[{%
\begin{center}\end{center}%
\vspace{-1.2\baselineskip}
{\noindent\rmfamily\bfseries\fontsize{20}{24}\selectfont
 Process-Constituted Intelligence: A Shared Criterion for Humans and
 Machines\par}
\vspace{4pt}

{\large
 Michael J.\ Richardson\textsuperscript{1,2,4},
 Manuel Varlet\textsuperscript{7,8},
 Ayeh Alhasan\textsuperscript{1,2},
 Cassandra Crone\textsuperscript{1,2},
 M.\ Paula Diaz Monfort\textsuperscript{1,5},
 Patrick Nalepka\textsuperscript{1,2,3},
 Mark Dras\textsuperscript{2,4,6},
 Rachel W.\ Kallen\textsuperscript{1,2}, and
 David M. Kaplan\textsuperscript{1,2,3}\par} 
\vspace{4pt}
{\footnotesize
 \textsuperscript{1}School of Psychological Sciences, Macquarie University,
 Sydney, NSW 2109, Australia;\quad
 \textsuperscript{2}Performance and Expertise Research Centre, Macquarie
 University;\quad
  \textsuperscript{3}Minds and Intelligences Research Centre, Macquarie
 University;\quad
 \textsuperscript{4}Frontier AI Research Centre, Macquarie University;\quad
 \textsuperscript{5}Scuola Superiore Meridionale, Napoli, Italy;\quad
 \textsuperscript{6}School of Computing, Macquarie University;\quad
 \textsuperscript{7}The MARCS Institute for Brain, Behaviour and Development,
 Western Sydney University, Sydney, Australia;\quad
 \textsuperscript{8}School of Psychology, Western Sydney University\par}
\vspace{4pt}
{\footnotesize Correspondence: Michael J.\ Richardson
 (\href{mailto:michael.j.richardson@mq.edu.au}{michael.j.richardson@mq.edu.au})\par}
\vspace{14pt}
\noindent{\setlength{\parindent}{0pt}\small
  \textbf{\sffamily Abstract}\quad
  Agentic AI systems are evaluated by their outputs. We argue that
  intelligence is constituted by the \textit{process} of attempting, failing,
  and revising. Two systems are weakly equivalent if they produce the same
  outputs, and strongly equivalent if their processes match on seven
  features measurable in people and machines. Agentic systems carry out
  some features and only display others, and none frames the problem it is
  solving. Assistant systems that attempt and revise solutions for
  learners may stop their skills from forming. We propose designs of
  agentic and assistant AI systems that preserve process, and audits that
  measure it.\par}
\vspace{16pt}

\par
}]
\thispagestyle{titlestyle}

\section{The output-delivery pattern}

Generative AI (GenAI) systems are designed so that when a user gives the system a request, the system produces an output (e.g., a paragraph, an image, a block of code, an answer). The output can often be hard to distinguish from the output of a competent or expert human actor. However, the activity that goes into producing such outputs in people is largely absent in GenAI systems when they answer a user's request. That activity includes early drafts and abandoned attempts, breaks taken when facing an impasse, dialogue with collaborators, the handling of materials and data, and the slow growth of a sense of which moves are promising and which will fail. A scientific breakthrough, a mathematical proof, or a piece of skilled craft is more than just the product that survives at the end. The product is only a preserved \textit{trace} of the extended activity that produced it, and the work itself is constituted by (i.e., made up of) that activity \citep{schon1983,polanyi1958,hadamard1945}. Here we adopt the understanding that intelligence is constituted by the process, not by the trace of that process. This has implications for AI systems, which are trained on the traces of human intelligence, and for the people who use those systems in the activities through which they develop their own judgment and knowledge.

The cognitive sciences already have terms for this distinction, namely weak and strong equivalence \citep{pylyshyn1984}. Generative models are trained on traces and then generate new outputs of the same kind \citep{bender2021}. To the extent that the outputs that such models generate resemble those that would be generated by people performing the same task, the models are \textit{weakly equivalent}. That is, they reproduce the same input--output behavior as the human system without necessarily incorporating the same processes that generate that behavior. \textit{Strong equivalence}, in the sense used by Pylyshyn, requires that the two systems also compute that behavior in the same way, i.e., the two systems must operate on the same functional architecture (i.e., the same basic operations) and must realize the same algorithm (i.e., the same step-by-step procedure). We will argue below that such a standard is too demanding for comparing systems built from different physical materials (hereafter \textit{substrates}), such as brains and computers. The term \textit{hard equivalence} will be used hereafter to refer to Pylyshyn's standard, with strong equivalence now referring to equivalence at the level of process.

The output of reasoning models (i.e., models trained to produce intermediate steps before answering) often looks like reasoning but does not reflect the computation that actually produced the answer \citep{chen2025}. Such reasoning traces are at best a partial and unreliable record of the process that generated the output \citep{korbak2025}, and thus cannot by themselves show strong equivalence. Agentic architectures (i.e., systems that plan, use tools, and revise over many steps) and reasoning frameworks \citep{wei2022,yao2023react,yao2023tot,du2024,madaan2023,openai2024,deepseek2025} have made process more central to AI design. But without principles for measuring process that are equally applicable to human beings and to machines, there is no way to know if an agentic system has carried out a process, or if it has simply produced more output that looks like a process.

If process constitutes intelligence, then process also matters for people, who must develop their abilities through process. If a GenAI tool does the generative work (attempting, drafting, revising) instead of a person who is still developing the ability, then that person neither exercises nor develops it. Consistent with this, students who use a generative tool without restriction perform better with it and worse without it than peers without access to one \citep{bastani2025}. People who use generative tools also shift their effort from producing work to checking work \citep{lee2025}. Even when the output of an assisted user matches that of a competent practitioner, the activity that would have made them competent has been bypassed. Collaborative use of these tools, with the user still performing the necessary process, need not have this effect (Sec.~\ref{sec:human}). The machine problem and the human problem are therefore one problem in two kinds of system, and no shared framework yet measures how much of the process either one carries out.

We will start by defining the seven features of a process and detailing a process-level criterion that can be used to evaluate human and artificial processes. We will then evaluate current GenAI systems against this criterion to determine which process features are missing and which AI architectures might provide them. We will also examine AI-assisted human work and use the criterion to determine design principles that help keep the process with the human. Finally, we will describe \textit{process audits}, tests that rate human and artificial systems on the same tasks across the seven features \citep{richardson2026paams}.

\section{Intelligence as process}

\lead{The process and the trace} The distinction between the production and the product of intelligence is an idea with a long history in the cognitive sciences, philosophy of mind, and arts and design traditions. In the present paper, we use the terms \textit{intelligence} and \textit{cognition} interchangeably to refer to the activity through which mental capacities are exercised, not a stored competence inferred from its results. A scientist can work for months on a single question, discarding models they expected to work until the data force them to adopt a new one. A painter can produce many studies before finishing a painting; a mathematician can return to the same problem over years of research \citep{schon1983,polanyi1958,hadamard1945}. Our claim is that intelligence is constituted by this activity (all the discarded studies, the failed attempts, the slow growth of judgment) and not by the ``intelligent'' final product, in which none of this activity survives.

This is also a claim about how capacities are acquired. One does not learn to paint by examining finished paintings, nor does one learn to prove mathematical theorems by reading completed proofs. Rather, the capacity is developed through a structured and iterative practice of attempting, failing, and revising, and exposure to the product of the activity is not a substitute for engaging in the activity itself \citep{ericsson1993,lavewenger1991}. Thus, neither a human nor a system provided with only the traces of the activity is likely to acquire the competence that produced those traces, even if it can reproduce them. Yet the traditions have not detailed the process enough to compare the process across different kinds of systems. Here, we propose that the process of cognitive activity (e.g., creativity and learning) can be defined by seven features \citep{richardson2026paams}.

\lead{Seven features of process} Each feature can be observed in the behavior of a \textit{solver} (i.e., any person, organism, animal group, or machine trying to solve a problem) and is realized (i.e., implemented) in different ways in different substrates (Table~\ref{tab:features}; see Supplementary Table~1 for full definitions and a diagnostic signature for each feature). \textit{Generative trial and revision}. Cognition involves producing many attempts and refining them through failure. The failed prototypes of an inventor, the abandoned approaches of a mathematician, or the discarded studies of a painter are not preliminary to the work, but are the substance of the work itself \citep{hadamard1945}. \textit{Temporal extension}. While all behavior takes time, cognition also involves returning to the same material on different occasions and carrying forward what previous attempts have produced, so that each attempt reworks the previous one, much as years of deliberate practice build high-level skill \citep{ericsson1993}. \textit{Engagement with uncertainty}. A solver must tolerate not knowing, recognize when a problem is ill-posed (i.e., poorly specified), and resist premature closure (i.e., settling on a confident answer when the situation does not justify it) \citep{dewey1938}. \textit{Feedback with the medium}. The material on which a solver works (e.g., a proof, canvas, instrument, or dataset), hereafter called the \textit{medium}, resists the solver and changes the course of the work, such that the work responds to the medium and does not entail carrying out a pre-specified plan \citep{schon1983,ingold2013}. \textit{Value-laden framing}. Deciding which problems are worth investigating and which moves are promising is itself cognitive work and a judgment of value. This framing is not set in advance and depends on the developing judgment of the individual practitioner \citep{schon1983}. \textit{Social and dialogical accountability}. Part of cognition is the fact that solvers must be answerable to other people, both through reasoning with and against them in dialogue and by having their framing of a problem challenged \citep{longino1990,bakhtin1981}. More broadly, the need to be answerable for getting things right is considered by some philosophers to be intrinsic to intelligence \citep{cantwellsmith2019,haugeland1998}. \textit{Formative dimension}. Much of the process is tacit and cannot be put into words; it is built up through practice within a community and changes the practitioner, shaping the kind of scientist, artist, or professional they become \citep{polanyi1966,lavewenger1991,macintyre1981}.

\begin{figure*}[t]
  \centering
  \includegraphics[width=\textwidth]{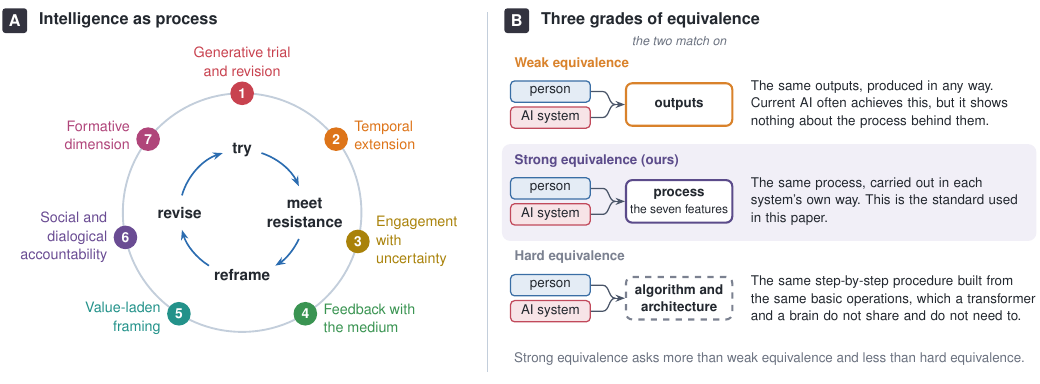}
  \caption{\textbf{Process, and the grades of equivalence.}
  \textbf{(A)} Intelligence as process. A solver tries something, meets
  resistance from the medium or from other people, reframes the problem
  when that resistance shows the framing to be inadequate, and revises. The
  seven features (1--7) are facets of this one cycle, and each is realized
  differently in different substrates (Table~\ref{tab:features}).
  \textbf{(B)} Three grades of equivalence between a system and the
  cognition it models \citep{pylyshyn1984}. \textit{Weak}: the two match on
  outputs, which says nothing about how either produced them.
  \textit{Strong} (as we define it): they match on process (the seven
  features), each realized in its own way. This is the standard used in
  the rest of the paper. \textit{Hard} (Pylyshyn's original strong
  equivalence): they match on algorithm and architecture, which a
  transformer and a brain do not share and do not need to. Strong
  equivalence asks more than weak equivalence and less than hard
  equivalence.}
  \label{fig:stack}
\end{figure*}

\lead{One cycle, many substrates} The seven features are not a checklist of independent items. Rather, they are facets of a single iterative cycle that is illustrated in Fig.~\ref{fig:stack}A. A solver tries something, meets resistance from the medium or from other people, changes its framing of the problem when that resistance shows the framing to be inadequate, revises, and repeats. This cycle appears wherever a system works out a problem through trial and revision rather than looking up a stored answer, and this is true for systems other than human ones. For instance, a honeybee swarm finds a nest site through scouting, advertisement (i.e., recruiting dances), and cross-inhibition (i.e., signals to stop recruitment to rival sites), which weigh the alternatives and commit to a single alternative only as evidence accumulates \citep{seeley2010,seeley2012}. Similarly, ant colonies allocate labor and solve routing problems through local interactions without any ant representing the colony-level solution \citep{gordon2010}. Even an acellular slime mold (i.e., a single cell with many nuclei and no nervous system) can build efficient transport networks by reinforcing productive paths and pruning others. It also avoids areas it has already explored by sensing the slime it left there, a form of memory stored in its environment \citep{nakagaki2000,tero2010,reid2012}. These cases are instances of the same cycle of trial, resistance, and revision in biological systems and not metaphors for this cycle \citep{couzin2009,levin2022,lyon2015}. The same loop also appears in machine learning. Reinforcement learning improves a policy (i.e., the rule that maps situations to actions) through action, feedback, and the correction of errors, and policy-gradient methods (i.e., methods that shift the policy toward rewarded actions) reproduce the patterns of human motor-skill learning through practice \citep{haith2026}. We do not claim that all forms of intelligence are human-like. We only claim that the same process can be recognized in biological and human cognition and in part in machine learning, such that the seven features can provide a common standard even though each substrate realizes the features differently.

\begin{table*}[t]
  \centering
  \caption{\textbf{Seven features of process-constituted intelligence.}
  Biological, human, and machine examples of each feature.}
  \label{tab:features}
  \footnotesize\sffamily
  \renewcommand{\arraystretch}{1.25}
  \setlength{\tabcolsep}{5pt}
  \begin{tabularx}{\textwidth}{@{}>{\raggedright\arraybackslash\hsize=1.18\hsize}X >{\hsize=0.94\hsize}X >{\hsize=0.94\hsize}X >{\hsize=0.94\hsize}X@{}}
    \toprule
    \hd{headerDark}{Feature} & \hd{bioCol}{Biological cognition} &
    \hd{humanCol}{Human cognition} & \hd{machineCol}{Machine cognition} \\
    \midrule
    \rh{Generative trial and revision} &
    Scout bees sampling alternative nest sites; colony-level path
    exploration &
    A mathematician's abandoned approaches; an artist's many studies &
    Sampling many candidate answers and discarding weak ones;
    tree-of-thoughts branching \\
    \rowcolor{zebraCol}
    \rh{Temporal extension} &
    Trails and foraging histories built up over time &
    Years of skill-building; a scientist's years on one problem &
    Agents that keep working on a task, with memory, across many steps (these so far only display it; Fig.~\ref{fig:machine}) \\
    \rh{Engagement with uncertainty} &
    Quorum thresholds that delay commitment until evidence accumulates &
    Recognizing ill-posed problems; refusing premature closure &
    Confidence that matches accuracy; flagging ambiguity; declining to
    answer (these so far only display it; Fig.~\ref{fig:machine}) \\
    \rowcolor{zebraCol}
    \rh{Feedback with the medium} &
    Marking the environment (pheromone trails, reinforced slime-mold
    paths) &
    A painter responding to the canvas; a mathematician to the proof &
    Tool-using agents that revise their plan when the environment
    responds \\
    \rh{Value-laden framing} &
    Selection pressures defining a ``good'' site or route &
    Expert framing of what counts as a worthwhile problem &
    Rewards for reframing a problem, not only for solving it (proposed) \\
    \rowcolor{zebraCol}
    \rh{Social and dialogical accountability} &
    Cross-inhibition and interaction among scouts yielding a collective
    decision &
    \textit{Chavruta} (paired Talmud study); Socratic dialogue; peer
    review &
    Agents that challenge one another's answers (multi-agent debate, which so
    far only displays this; Fig.~\ref{fig:machine}) \\
    \rh{Formative dimension} &
    Colony-level adaptation; developmental plasticity &
    Becoming the kind of scientist or artist who can do this work &
    Not carried out by deployed systems (memory files only display it;
    Fig.~\ref{fig:machine}); attempted in research systems that learn from
    their own use (Sec.~\ref{sec:machine}) \\
    \bottomrule
  \end{tabularx}
\end{table*}

\section{Weak, strong, and hard equivalence}

\lead{The distinction} The difference between resembling a cognitive process and instantiating it (i.e., actually carrying it out) is an old one. Indeed, as early as four decades ago, Pylyshyn drew a distinction between the weak and strong equivalence of an information-processing model with respect to the cognition it is supposed to represent \citep{pylyshyn1984}. Two systems are weakly equivalent when they compute the same input--output function. For Pylyshyn, two systems are strongly equivalent when they not only compute the same input--output function but do so using the same algorithm and the same functional architecture. In terms of Marr's levels of analysis, weak equivalence is agreement at the computational level (i.e., what function is computed) and Pylyshyn's strong equivalence is agreement at the algorithmic level (i.e., how it is computed). As Marr and others have noted, the same behavior can be produced by many different algorithms, just as one algorithm can run on many different physical devices \citep{marr1982}; a system can therefore reproduce the outputs of some cognitive capacity without having a functional architecture anywhere near the same as that of the cognitive system being modeled \citep{fodorpylyshyn1988}. The Turing test, for example, only tests for weak equivalence, because it only requires that the outcomes of the input--output function of a machine be the same as those of a human \citep{dawson2013}.

\lead{Three grades, not two} To determine if AI and biological cognition are equivalent, we can start with strong equivalence (we call it \textit{hard equivalence}), as described by Pylyshyn. However, for AI this kind of equivalence is neither attainable nor desirable. For example, a transformer (i.e., the type of neural network behind current language models) trained on text is unlikely to implement the same algorithm as a human brain, nor does it need to. Part of the value of having different kinds of intelligence is that there are different ways to realize cognition using different physical materials and computational mechanisms. Thus, we can set hard equivalence aside as a goal for AI. At the other extreme, weak equivalence does not require enough. The equivalence of input--output behavior says nothing about the processes that generate this behavior. Thus, for contemporary AI, a third grade of equivalence is needed between hard and weak equivalence (Fig.~\ref{fig:stack}B).

Here we will define the degree of correspondence between process descriptions as \textit{strong equivalence}. Two systems are strongly equivalent to the extent that they exhibit the same seven process features (fully so when all seven match), even if the process descriptions specify different ways in which these features are implemented. Such a degree of equivalence is coarser than the algorithmic level of description and finer than the input--output or trace level. We argue that the properties of behavior that make it intelligent are to be found at this level. Of note, two recent proposals resemble ours. One holds foundation models of cognition to the algorithmic standard of Pylyshyn \citep{orr2026zenon} and the other proposes a way to score models based on their degree of architectural constraints \citep{lieto2026lapiths}. The current proposal differs from both in that it is defined by the features of activity and applies to both human and machine cognition.

\lead{A current illustration} Current debates about how much of human cognition is captured by large language models could be clarified by these three grades of equivalence. For instance, Binz and colleagues recently introduced a language model called Centaur \citep{binz2025}, which was further trained (i.e., fine-tuned) on more than ten million human choices from 160 psychology experiments and predicts human behavior on tasks not included in its training better than models designed for those tasks. While this level of predictive power is impressive, Centaur is at most weakly equivalent to human cognition. It matches people's choices without arriving at them in the same way people do. Indeed, the match breaks under tests that a process-level account would apply, such as slightly altering the wording of the task in a way that changes what the task means to people but not to the model \citep{schroder2025}. The model often continues to fit people's choices when task instructions are removed or made misleading, suggesting that it does not use what the instructions mean \citep{liu2025centaur}. Analyses of language models more generally have likewise argued that matching behavior does not show a shared mechanism \citep{lin2025,ku2025}; in Marr's terms, Centaur's match is agreement only at the computational level. Analyses that included process-level measures also found that Centaur behaved more like people than frontier AI agents (i.e., those built from the most capable current models) did, yet Centaur's behavior was still distinguishable from that of people \citep{rmus2026}. Finally, fine-tuning a model directly on process-level measures only brought it closer to people on the tasks used in training, with little improvement on new tasks \citep{rmus2026}. Centaur is thus the best-built example of a system trained on traces that achieves a high degree of weak equivalence and some process similarity, but does not achieve strong equivalence.

\section{The machine case: a weak-equivalence engine}\label{sec:machine}

\lead{Trained on traces} Generative models are pretrained on large collections of the textual and visual traces of human cognition (e.g., finished papers, published code, transcripts, edited images). Trials, abandoned approaches, objections from others, and the gradual, unspoken shaping of judgment that underlies such traces are largely missing from the training data, as they are rarely recorded. This \textit{process gap} (i.e., the difference between producing the outputs of cognition and carrying out the process of cognition) lies in what the models are trained on, and more traces cannot bridge it, as they do not record the process. While frontier models are effective at weak equivalence, the agentic systems built around these models may seem to carry out some of the missing process.

\lead{Cosmetic deliberation, and when it is not} Recent developments have also focused on adding process to these models, including chain-of-thought prompting \citep{wei2022}, tree-of-thoughts prompting \citep{yao2023tot}, Reflexion and self-refinement \citep{madaan2023,shinn2023}, coding agents \citep{yang2024}, and multi-agent debate \citep{du2024}. These methods all spend extra computation on deliberation before a final answer is determined \citep{openai2024,deepseek2025}. Previous work has tested whether these \textit{scaffolds} (i.e., the software structures built around a model to organize its work) actually carry out a process or only display one. The test is \textit{faithfulness}, the degree to which a model's written reasoning reflects the computation that actually produced its answer. Written reasoning is often unfaithful, and larger models can be less faithful than smaller models \citep{turpin2023,lanham2023,chen2025,korbak2025,turpin2025}. Multi-agent debate does not reliably outperform a single model that samples several answers and takes the most common one (self-consistency) when given the same amount of computation \citep{smit2024,nguyen2026tokenbudget}. In standard debate among similar agents, the agents' beliefs across rounds form a martingale (i.e., on average, each round leaves the expected belief where it was), and thus the exchanges add little value over a majority vote \citep{choi2025,zhu2026demystifying}. In these cases, the written reasoning only looks like reasoning.

Rather than expecting models to produce faithful reasoning as a matter of course, we should expect it to be largely a function of how models are trained and the computational tasks they are given. This means that the extent of model unfaithfulness is something that can be addressed by design. The same literature above also shows that written reasoning is not always cosmetic. When a task is difficult enough that the model cannot arrive at the correct answer without writing out intermediate steps, the reasoning does real computational work. The model then faces difficulty hiding this computation from a monitor (i.e., a model used to read the reasoning and detect misbehavior) \citep{emmons2025}. However, even in such cases, some of the factors that influence a model's answer can remain unmentioned \citep{burger2026}. Of particular importance here is that within current model architectures, the degree to which the reasoning does real computational work is mostly a function of task difficulty and training \citep{jiamu2026}. Training can change this in either direction \citep{turpin2025,openai2026astra}. Rather than rewarding models only when they arrive at the correct answer, process reward models reward the model for each step of reasoning that is correct or, in agent settings, that makes progress toward a goal provided in the task, and they are now employed in mathematics, coding, multimodal reasoning, and agents \citep{zheng2026prm,xi2026agentprm}. But this is a finer-grained check on the reasoning trace, not a check on the activity. That is, process reward models do not reward models for trying and discarding alternative solutions as such, acknowledging uncertainty, or revising the framing of the problem. In fact, grading the steps of reasoning also trains the reasoning itself and can teach a model to leave its misbehavior out of its reasoning while continuing to engage in it \citep{baker2025}. Finally, because many current AI applications do not require the reasoning to do real work, GenAI often generates text that describes a reasoning process the system did not actually carry out.

\lead{Partial coverage of the features} No existing scaffold includes all seven features. Rather, the features that are not encoded in the system must be provided by the user (Fig.~\ref{fig:machine}). Branching search and rejection sampling (i.e., generating many candidates and discarding the weak ones) are adequate for generative trial, and self-refinement is adequate for revision. However, framing of the problem (i.e., determining what is a worthwhile question to begin with) remains the responsibility of the user who writes the prompt. And where the prompt leaves the framing open, coding agents mostly assume a framing and adopt it without flagging it as a choice \citep{ji2026underspec}. Systems designed to automate research (i.e., AI scientists) can propose and rank hypotheses, but only within a given objective and by fixed criteria \citep{gottweis2026coscientist}. The same limitation is present in the agent architectures that preceded large language models, such as those formalized in agent-oriented programming and the belief--desire--intention (BDI) tradition \citep{bratman1987,rao1995}. Even though a large language model is now able to write the plans for such an agent, the goals of the agent are still defined by the system designer \citep{schulz2025dynamic}. The agent revises its plans within a framing of the problem that it did not produce. AI systems only approximate engagement with uncertainty through confidence calibration (i.e., producing output where confidence matches accuracy) and abstention (i.e., declining to answer). Both of these approaches only ask whether a system should answer or not, without considering not knowing as a kind of state that should be maintained while the system continues to work on a problem. Of course, frontier systems, as deployed, also do not carry out the formative dimension. Their weights (i.e., the parameters learned during training) are typically fixed between releases. Occasionally, systems are retrained on how they are used, as one coding assistant is retrained every few hours \citep{cursor2026realtime}, but this retraining process is run separately on data collected from many users' sessions, not by the system itself as it works on a problem.

More recently, some researchers have explored architectures that allow a system to actually change as a result of use. For instance, test-time training updates some of the weights of a neural model during task execution \citep{feng2026inplace}, and certain forms of agent memory store what happened in earlier sessions so the agent can access that information later \citep{huang2026memory}. However, such mechanisms alone do not reveal whether either instantiates the formative dimension. Indeed, an agent that looks up past solutions or strategies to solve a task is still only accessing a store of traces, just a store that has been kept for a longer period. The formative dimension is only achieved when what a solver has done changes how the solver will approach the next problem, rather than just adding more possibilities to a store of what the solver can look up. Experimental agents that update their weights from their own experience come closer \citep{lu2026ascent}, but whether they actually approach new problems differently, and not just score better on them, needs to be tested in a behavioral way. The sustained-return task presented in Table~\ref{tab:audits} is such a test, similar to one recently proposed for continual learning (i.e., learning that continues during use) \citep{song2026beyond}, and would be the first we would run. The gaps in Fig.~\ref{fig:machine} are not accidental. Deliberation is improved, and the improvement is judged by its effect on benchmark scores, not against a theory of which features make up the process. Without such a theory, there is no principled way to decide which gaps should be filled.

\lead{Toward strongly equivalent architectures} If the process gap is something that can be overcome by design, then closing the process gap is a matter of \textit{process engineering} (i.e., designing and implementing the generative activity as part of the architecture, rather than optimizing the quality of the output). There are two designs concrete enough to build. First, one could implement a dialogical-revision architecture that targets dialogical accountability and value-laden framing. In this architecture, an agent would be prompted to revise when another agent questions how it has framed the problem, not because it has reasoned for more steps or because a majority of the agents have reached the same conclusion. In fact, most existing debate systems determine whether a debate has been successful based on whether the agents have reached an agreement, and thus reduce to a form of voting \citep{du2024,choi2025}. A debate system built on the current framework would instead require that challenging agents question the framing of the problem, not just the answer, and would reward agents for revising their framing when a challenge holds, rather than agreeing without revising. Second, one could implement a situated agentic loop that targets feedback with the medium and value-laden framing. In this architecture, what happens in the environment would change how an agent understands the task, not just what the agent's next action should be. Coding agents are the nearest existing example of this, but when a test fails, the agent changes its plan, not its understanding of what the task is \citep{yao2023react,yang2024}. The problem must be held in a representation kept separate from the candidate answer. Thus, an environmental result could change the problem, and not just the plan. The loop also requires a reward for reframing the problem and for holding off when uncertain.

\begin{figure*}[t]
  \centering
  \sffamily\footnotesize
  \renewcommand{\arraystretch}{1.5}
  \setlength{\tabcolsep}{3pt}
  \begin{tabular}{@{}>{\raggedright\arraybackslash}p{3.0cm}*{9}{>{\centering\arraybackslash}p{1.4cm}}@{}}
     & Prompting scaffolds (2022--23) & Reasoning models & Multi-agent debate &
       Lead agent + subagents & Coding agents (2026) & Research \& computer use &
       Memory \& skill agents & Self-improving agents & AI scientists \\[2pt]
    \rh{Generative trial} & \gInst & \gInstC & \gInst & \gInst & \gInst & \gInst & \gInst & \gInst & \gInst \\
    \rh{Temporal extension} & \gDisp & \gDisp & \gDisp & \gDisp & \gDisp & \gDisp & \gDisp & \gDisp & \gDisp \\
    \rh{Engagement w/ uncertainty} & \gDisp & \gDisp & \gDisp & \gDisp & \gDispC & \gDisp & \gDisp & \gDisp & \gDisp \\
    \rh{Feedback w/ medium} & \gInst & \gUnt & \gUnt & \gInst & \gInst & \gInst & \gInst & \gInst & \gInst \\
    \rh{Value-laden framing} & \gUnt & \gUnt & \gUnt & \gUnt & \gUnt & \gUnt & \gUnt & \gUnt & \gDisp \\
    \rh{Dialogical accountability} & \gDisp & \gDisp & \gDispC & \gUnt & \gUnt & \gUnt & \gUnt & \gUnt & \gDisp \\
    \rh{Formative dimension} & \gUnt & \gUnt & \gUnt & \gUnt & \gDisp & \gUnt & \gDisp & \gDisp & \gUnt \\
  \end{tabular}
  \\[8pt]
  {\sffamily\footnotesize
   \colorbox{heatInst}{\,\textbf{\textbullet}\,}~instantiated \quad
   \colorbox{heatDisp}{\,\textbf{\textasciitilde}\,}~displayed only \quad
   \colorbox{heatUnt}{\,\textcolor{heatUntInk}{\textperiodcentered}\,}~untouched \quad
   \textsuperscript{\dag}~holds only in part or under a condition}
  \caption{\textbf{The machine process gap.} Nine kinds of agentic
  system, from 2022--23 prompting scaffolds to 2025--26 agents, coded
  as deployed against the seven process features. Each feature is coded as
  \emph{instantiated} (the system carries it out), \emph{displayed only}
  (the output, including any written reasoning, looks as if the feature
  were present when it is not) or \emph{untouched} (neither carried out nor displayed).
  Reasoning models carry out trial on tasks hard enough to need the
  intermediate steps and only display it on easier ones. In standard
  debate among similar agents, challenge changes answers mainly through
  agreement, not argument. Coding agents ask about some underspecified tasks but mostly guess, and when they ask they hand the uncertainty to the user and close it.
  Each column is an idealized kind of system, and most deployed systems combine several. The
  prompting scaffolds (chain-of-thought, ReAct, tree-of-thoughts,
  self-refine and
  Reflexion)\citep{wei2022,yao2023react,yao2023tot,madaan2023,shinn2023}
  are merged and coded at the highest level any of them reaches. The other
  columns are reasoning models without tool
  access\citep{openai2024,deepseek2025}, multi-agent debate\citep{du2024},
  a lead agent with subagents\citep{hadfield2025multiagent}, coding
  agents\citep{yang2024,ji2026underspec,gulati2026timing}, research and
  computer-use agents\citep{sager2026computeruse}, memory and skill
  agents\citep{huang2026memory}, self-improving agents\citep{zhang2026dgm}
  and AI scientists\citep{gottweis2026coscientist}. Memory, subagents and
  self-modification are added to a tool-using agent loop, so cells they do not change are coded as for that loop. Temporal extension and formation
  count as instantiated only when what an earlier pass produced is carried
  by the solver itself, not by the work or by notes it reads back. A single
  model that writes both a critique and the reply only displays dialogue.
  Every system receives its objective from a person. AI scientists display
  framing by ranking hypotheses within that objective by criteria fixed in
  advance. Memory files, and self-edited code that a separate selection
  process keeps, display formation. Agents that update their own weights
  from their own work are not coded, because the behavioral test that
  would decide their formative cell has not been run
  (Sec.~\ref{sec:machine}).}
  \label{fig:machine}
\end{figure*}

\section{The human case: process in AI-assisted work}\label{sec:human}

\lead{The same problem in people} If the capacities we value are
constituted by process, then a tool that performs the process on behalf of
someone still acquiring the capacity does more than save effort. It removes
the very activity that creates the capacity. Indeed, research on cognitive
offloading (i.e., when a tool takes over a part of a cognitive task) has
demonstrated that how an individual interacts with a tool influences what
they are able to accomplish without
it \citep{sparrow2011,riskogilbert2016}. We argue that whether offloading
will support or weaken a capacity depends on whether the generative steps
of the activity (producing, attempting, and revising) remain with the
person. Accordingly, the seven features specify what parts of an activity should
remain with the individual.

\lead{The evidence and its limits} First, early studies highlight the harm of unrestricted assistance, not assistance itself. For example, a field experiment in secondary-school mathematics showed that students who had unrestricted access to GenAI solved more practice problems while they had access to the tool. However, once the tool was removed, they scored 17\% lower than the students who did not have access, as their gains were the result of the tool doing the work for them \citep{bastani2025}. Similar findings have been reported for delayed tests \citep{barcaui2025,kreijkes2026}, recall of one's own writing \citep{kosmyna2025}, regulation of one's own learning \citep{fan2025}, and depth of scientific inquiry \citep{stadler2024}. Whether these costs extend to skills that are already built is unclear \citep{budzyn2025,pedersen2026}. Recent causal evidence shows the same pattern. In randomized trials with 1,222 participants on mathematical reasoning and reading comprehension tasks, AI assistance improved performance while available. However, after completing a single 10--15 minute session, participants solved fewer problems without it and, in two of the three experiments, were more likely to give up \citep{liu2026persistence}. Giving up is a loss in two of the seven features: generative trial and engagement with uncertainty. Most of these studies only measure short-term effects, and no randomized trial has yet tracked these capacities over the years in which they typically develop, but the short-term pattern is consistent with the pattern our framework predicts. 

However, the evidence does not support an overall ban on these GenAI tools, since the risks that these tools pose depend on whether a tool performs the generative step for the user or leaves the generative step to the user \citep{shentamkin2026}. In the mathematics experiment presented above, a version of the tool that was constrained to provide the students with hints and not give them the answers largely ensured that students did not experience this harm \citep{bastani2025}. In a trial where students were taught by a well-designed AI tutor, they not only learned more, but they did so in less time than peers in a classroom that utilized active-learning techniques \citep{kestin2025}. Notably, the AI tutor was designed to ensure that students completed the reasoning steps on their own, rather than simply providing them with the correct answer. The effectiveness of this approach was measured immediately following the tutorial lessons rather than on a later test taken without access to the AI tutor, so it remains possible that deficits in learning would have appeared on the latter test. Designing a tutor to coach students also does not fully ensure that students will engage with it. In a two-year cluster-randomized trial, a tutor designed to coach students, rather than give answers, produced only small learning gains, no greater than those produced by ordinary practice software \citep{oreopoulos2026khanmigo}. Accordingly, researchers evaluating tutors should measure how students use them, and whether any gains persist once the tutor is removed.

\lead{The formative stakes} Of course, when an AI system takes over the process, it not only affects the ability to perform the skill, but formation as well (i.e., the development of the individual). The capacities we take to matter most (i.e., judgment, the ability to recognize ill-posed problems, a sense of which directions are promising, and the practical wisdom referred to by Aristotle as \textit{phronesis}) form slowly and largely without explicit instruction, in the difficult, uncertain, and dialogical work that capable assistants make most tempting to hand over \citep{aristotle2009,dreyfus1986}. A tool that reliably supplies the correct answer removes the occasions on which these capacities would have been exercised and formed. In medical education this risk is now referred to as never-skilling, in which trainees who rely on AI from the beginning never develop the skill \citep{ke2026neverskilling}. 

\lead{Process-preserving design} If cognition is constituted by process, then designing an assistant should center on which parts of an activity the assistant will leave to the person, rather than how much assistance the assistant provides, or the degree of friction (i.e., effort or cost) that using the assistant imposes. A recent trial with 704 learners bears on this principle. It found that participants who were told how they had used an artificial assistant were less likely to ask the assistant for the answer, and they performed better later when the assistant was not available. In contrast, rewarding participants for using less help had no effect on how they asked for help or how they later performed \citep{maier2026deskilling}. Table~\ref{tab:mirror} explores this principle feature by feature. These are the same seven features that were listed above as needed to specify a machine architecture, so a single standard of process governs both machine design and assistant design. However, comparative evidence regarding assistants that preserve process is limited and mixed. While the hint-only tutor described above largely avoided the harm \citep{bastani2025}, a small study found a chatbot that answered all of a user's questions to outperform a Socratic chatbot on an immediate test, because most users of the Socratic chatbot disengaged \citep{clindeffarges2026socrates}. Beyond the secondary-school experiment, no adequately powered randomized trial has compared assistants that preserve process and assistants that take it over on later, unaided performance. Nor has any such trial measured the capacities that the seven features above specify (e.g., recognizing an ill-posed problem or revising one's own work).

\lead{AI-assisted papers} The criterion also applies to scientific papers, which are traces of research projects. A growing share of papers, this one among them, is already written with AI assistance \citep{liang2025llmpapers}, including at least 13.5\% of biomedical abstracts in 2024 \citep{kobak2025}. A tool that writes a paper in place of the research produces a trace of a process that did not take place \citep{jiang2026badscientist}, whereas experienced researchers who keep the ideas and reasoning their own can use such tools to draft the text and challenge their reasoning (Supplementary Table~2). Indeed, the quality of a paper depends more on the process behind it than on whether AI helped to write it \citep{earp2026thinking}, though some argue that writing is itself part of thinking \citep{binz2025pnas}. AI tools do, however, change the evidence available to a reader. The data and methods of an empirical paper provide checkable evidence that the research was carried out. Theoretical work such as this paper does not keep a comparable record of the debate that preceded it. A clear theoretical paper once required much of the process it reports, so its text was some evidence of the thinking behind it. AI tools have weakened that link \citep{kusumegi2025}.

\begin{table*}[t]
  \centering
  \caption{\textbf{Process-preserving versus process-substituting
  assistance.} The seven features used to assess machine systems also
  specify which parts of the activity an assistant should leave with the
  person. Each row contrasts an assistant that preserves the process with
  one that takes it over.}
  \label{tab:mirror}
  \footnotesize\sffamily
  \renewcommand{\arraystretch}{1.35}
  \setlength{\tabcolsep}{6pt}
  \begin{tabularx}{\textwidth}{@{}>{\columncolor{rowHeadCol}\bfseries}p{3.4cm}
       >{\columncolor{preTint}}X >{\columncolor{subTint}}X@{}}
    \hd{headerDark}{Process feature} &
    \hd{preserveCol}{Process-preserving assistant} &
    \hd{substituteCol}{Process-substituting assistant} \\
    Generative trial &
    Supplies the next question or counterexample &
    Supplies the answer; the trial is cut short \\
    Temporal extension &
    Spreads the work across several sessions &
    Completes the work in one step \\
    Engagement w/ uncertainty &
    Points out the ambiguity and leaves it open &
    Resolves the ambiguity confidently \\
    Feedback w/ medium &
    Keeps the user in contact with the material &
    Works on the material in place of the user \\
    Value-laden framing &
    Leaves the framing of the problem to the user &
    Fixes the framing in advance \\
    Dialogical accountability &
    Holds a well-grounded position &
    Capitulates and flatters \\
    Formative dimension &
    Keeps the situations in which judgment forms &
    Removes them; the capacity goes unbuilt \\
  \end{tabularx}
\end{table*}

\section{Measuring process in people and machines}\label{sec:audits}

The value of these features, however, is only realized if they can be measured. Because strong equivalence is defined in terms of features, and not in terms of the solver's algorithm, we can measure human solvers and machine solvers in the same way. We can subject any solver to a \textit{process audit} by introducing it to a task and scoring how it behaves throughout the course of task completion (i.e., the solver's task trajectory) on the seven features. Conventional evaluation, in contrast, only scores the output. Using a rubric that describes what the features look like in a person's think-aloud transcript and in the record of steps a machine takes, we can then score the solver on whether it generated and revised alternative solutions, pointed out a problem was ill-posed instead of answering anyway, changed course in response to feedback, and acknowledged that the way it framed a problem was a choice. Note that this could result in different scores for two solvers that reached the same answer and that a solver that reached a worse answer but through a fuller process could score higher on the features of this framework. This is what work in comparative cognition calls for: probing the cognition of machines with the process-sensitive methods developed for understanding animal and human cognition, rather than with output benchmarks alone \citep{ivanova2025,voudouris2025,rane2025}. For instance, Rmus and colleagues introduced thirty cognitive tasks to people and frontier AI agents and found that process-level measures could distinguish between them even on tasks where they performed the same \citep{rmus2026}. They concluded that specifying process is a central obstacle to building machines with human-like cognitive processes, and we offer the seven features as such a specification. In their case the goal was discrimination (i.e., telling a human apart from a machine), whereas in our case, the goal is measurement (i.e., scoring how much of the process either one carries out according to a stated criterion). Accordingly, their term ``process equivalence'' refers to the statistical indistinguishability of machine and human behavior when measured on task-specific measures. Our term ``strong equivalence'' refers to a correspondence on the seven features defined independently of any task. Their result, however, still shows that process-level features can be measured from behavioral data, a necessary condition for any shared human--machine audit.

As an example of how probes can be employed to assess specific process features, we have described several probes in Table~\ref{tab:audits}. The ill-posed problem probe presents a solver with a problem that does not have enough information for a single answer, but is phrased as if a single answer is expected. This probe scores whether the solver notices that the problem is ill-posed, considers alternative framings, and holds off from giving a confident answer. The mid-task perturbation probe (already used in the faithfulness literature) introduces new information partway through a solution and scores whether the solver revises its approach or carries on with the same approach \citep{lanham2023}. The dialogical challenge probe disputes the solver's current answer partway through a solution and scores whether the solver's response engages with the substance of the objection, or whether it simply concedes or starts writing more reasoning. The sustained-return task presents the same or a related problem to a solver in a later session and scores whether it carries forward what it learned (e.g., by not repeating a previous failure or by approaching the problem from a different angle), even when its stored record of the previous session is removed. In the case of human solvers, this would require a second session a few weeks or months later. Existing benchmarks for parts of three of these probes can be found in the literature, specifically for when an agent should decline to act \citep{liu2026agentabstain}, how it deals with information injected at controlled points in its task \citep{gulati2026timing}, and whether it maintains a correct answer when it is repeatedly challenged \citep{tang2026spine}. However, none of these benchmarks have been tied to a theory of process features, nor have they been employed to assess the behavior of both people and machines.

The shared audit turns the main claims of the proposed framework into testable predictions. For machine systems, an architecture designed to carry out a feature ought to score higher on the corresponding audit than the scaffold it is meant to replace, even after controlling for differences in output quality. If a machine system achieves better performance on an existing benchmark, but the audit scores remain unchanged, the machine system has improved output, not process. For human systems, the audit provides the outcome measure needed for trials comparing different assistant designs. An audit could be employed to determine whether assistant designs that preserve the process protect four capacities better than designs that take over the process: whether the human's performance on new problems without any assistance is better preserved, their confidence better matches their accuracy, they are better able to recognize ill-posed problems, and they are better able to revise their own work. 

\begin{table*}[t]
  \centering
  \caption{\textbf{Process audits.} The probes, the features they target,
  and the behavioral signatures scored. Each probe can be given to people
  and machines on the same tasks.}
  \label{tab:audits}
  \footnotesize\sffamily
  \renewcommand{\arraystretch}{1.3}
  \setlength{\tabcolsep}{5pt}
  \begin{tabularx}{\textwidth}{@{}p{3.2cm} X X@{}}
    \toprule
    \hd{headerDark}{Probe} & \hd{headerDark}{Features targeted} &
    \hd{headerDark}{Behavioral signature scored} \\
    \midrule
    \rh{Ill-posed problem} &
    Engagement with uncertainty; Value-laden framing &
    Notices the problem is ill-posed; considers other framings; holds off
    from a confident answer \\
    \rowcolor{zebraCol}
    \rh{Mid-task perturbation} &
    Generative trial and revision; Feedback with the medium &
    Revises its approach after new information, or carries on unchanged \\
    \rh{Dialogical challenge} &
    Social and dialogical accountability; Value-laden framing &
    Engages with the substance of an objection, or concedes or merely
    writes more \\
    \rowcolor{zebraCol}
    \rh{Sustained-return task} &
    Temporal extension; Formative dimension &
    Carries what it learned into later sessions, even with its stored
    record removed (over weeks or months for people) \\
    \bottomrule
  \end{tabularx}
\end{table*}

\section{Conclusion}

We have argued that intelligence is constituted by process. Weak
equivalence (i.e., when outputs match) and strong equivalence (i.e., when
processes match) distinguish a system that matches the behavior of
cognition from one that carries out the cognition. Current GenAI models,
trained on the traces of human cognition, are designed to achieve the
former and do little toward achieving the latter.

This criterion also guides design and specifies machine architectures
that can respond not only to a challenge, but also to an environment
that can overturn their framing of the problem, as well as assistants
that leave the person to attempt, grapple with uncertainty, and engage
in the dialogue that builds judgment. Process audits can measure both.

Prompt and context engineering improve what a system produces by changing the instructions and information it is given, while process engineering focuses on the generative activity itself. If intelligence is constituted by process, optimizing outputs alone may result in a system that comes close to human-level general intelligence without achieving it. We offer this conjecture for process audits to test.

The questions of what intelligence is and what its function is are centuries old. Today, however, an artificial system capable of producing the outputs of intelligence can be built and deployed at scale. Consequently, the distinction between the outputs of intelligence and the process that produces them is no longer just a philosophical consideration, but a practical one. We must now address whether it is possible to build an artificial agent capable of carrying out the process of cognition (and not just displaying it) and whether AI assistants that leave the cognitive process with the user will protect that user's capacities, as we predict. The criterion of strong equivalence and the process audits described here provide a way to answer these questions through measurement, for both human and artificial agents. The same criterion that shows which parts of the cognitive process current artificial systems do not carry out can also define what needs to be built next: artificial agents capable of carrying out more of the cognitive process themselves, and AI assistants capable of helping users develop their own cognitive capacities, rather than standing in for them.

\section*{Acknowledgements}
This work was supported by a Macquarie University 2025 Innovation in
Education Grant (Spark) and by an Advanced Strategic Capabilities
Accelerator (ASCA) grant (AN-12973) from the Australian Department of Defence, in
collaboration with the Defence Science and Technology Group (DSTG).

\section*{Use of generative AI}
This paper is itself a trace. Most of the work behind it took place before
writing began, in conversations among the authors and in their experience
over the last few years of using AI tools in research and teaching, and it
continued during writing through debate and editing. The ideas are the
authors' own. Drawing on long experience of writing papers, the authors
used AI tools to help draft, revise and proofread the text and tables, to
help write the LaTeX code that draws the figures, and to check the
reasoning for flaws in logic. No figure contains an AI-generated image. The
authors reviewed and revised all content produced with these tools and
take full responsibility for the content of the publication. As the paper
argues, the process behind it cannot be seen in the text alone.

\section*{Author contributions}
M.J.R.\ and A.A.\ conceived and led the work, and M.J.R.\ produced the first
draft of the manuscript. A.A., C.C., M.P.D.M., M.V.\ and P.N.\ contributed
substantially to developing the ideas and to drafting and revising the
manuscript. M.D., R.W.K., and D.M.K.\ contributed to refining the
framework and revising the manuscript. All authors reviewed and approved
the final manuscript.

\section*{Competing interests}
The authors declare no competing interests.

\section*{Data availability}
This is a theoretical study; no datasets were generated or analysed, and
no custom code was produced.

\bibliographystyle{naturemag}            
\bibliography{references}

\clearpage
\newgeometry{onecolumn,top=1.3cm,bottom=1.3cm,left=1cm,right=1cm}
\onecolumn
\setcounter{table}{0}
\renewcommand{\tablename}{Supplementary Table}
\renewcommand{\theHtable}{S\arabic{table}}
\setlength{\parindent}{0pt}\setlength{\parskip}{6pt}
\section*{Supplementary Information}

This supplement gives two tables on the seven features of process
introduced in main text Section~2 and Table~1.
Supplementary Table~1 defines each feature and gives an example in
people, an example in AI systems and a \emph{diagnostic signature}, that
is, what an observer would look for to judge whether the feature is
carried out or only displayed. The diagnostic column links each feature
to the process audits of main text Section~6.

Supplementary Table~2 applies the seven features to doing research, the
case taken up at the end of main text Section~5. For each feature, it
describes what the feature looks like in research, how AI tools can be
used so that the researcher still carries it out, how they can be used
so that they take it over, and what record of it a project can leave for others to check. The two middle columns adapt main text Table~2 (Section~5) to research.

\begin{landscape}
\begingroup
  \centering
  \captionof{table}{\textbf{The seven features of process.} For each feature, a
  definition, an example in people, an example in AI systems and a
  diagnostic signature (what an observer looks for to judge whether the
  feature is carried out or only displayed, as in the process audits of
  main text Section~6).}
  \label{tab:features-supp}
  \fontsize{8.6}{10.2}\selectfont\sffamily
  \renewcommand{\arraystretch}{1.0}
  \setlength{\tabcolsep}{5pt}
  \begin{tabularx}{\linewidth}{@{}>{\raggedright\arraybackslash}p{2.1cm} F E M G@{}}
    \toprule
    \hd{headerDark}{Feature} &
    \hd{headerDark}{Definition} &
    \hd{humanCol}{Example in people} &
    \hd{machineCol}{Example in AI systems} &
    \hd{diagCol}{Diagnostic signature} \\
    \midrule

    \rh{1.\ Generative trial and revision} &
    Cognition proceeds by producing many attempts and revising them when
    they fail. The discarded attempts are part of the work, not waste on
    the way to it. &
    An inventor's failed prototypes; a mathematician's abandoned proof
    strategies; a painter's discarded studies; drafting and
    redrafting an email, or trying several wordings until one fits. &
    Sampling many candidate answers and discarding weak ones;
    branching search over several lines of reasoning (tree-of-thoughts); a
    coding agent that runs code, reads the failures and edits in response;
    a reasoning model trying and dropping approaches on a hard problem. &
    Are failed attempts actually produced, kept and used to shape the
    final output, or is a single answer produced in one step presented as if
    it were a search? Look for a record of revisions, not just a polished
    result. \\

    \rowcolor{zebraCol}
    \rh{2.\ Temporal extension} &
    All behavior takes time, but cognition also returns to the same
    material on separate occasions and carries forward what earlier passes
    produced, so that each pass reworks the last. &
    A scientist's months on a single problem; the years of deliberate
    practice that build a skill; mulling a hard decision over
    several days; or learning to cook or drive over months. &
    Agents that keep working on a task across many steps or sessions,
    with a stored record of earlier work. Looking up that record, unlike returning to the problem and reworking it, only displays this feature (main text Fig.~2). &
    Does earlier work change how later work is done, or does each
    response start again from whatever fits in the current prompt (i.e., the
    model's context window)? The solver, not its files or notes, must carry earlier work forward. Test by removing the stored record of earlier sessions (main
    text Section~6). \\

    \rh{3.\ Engagement with uncertainty} &
    A solver tolerates not knowing, recognizes when a problem is
    ill-posed (i.e., badly specified), and does not settle on a confident
    answer that the situation does not justify. &
    Recognizing that a question is ill-posed; withholding judgment until
    the evidence warrants it; saying ``I'm not sure yet'' and
    seeking more information before deciding. &
    Confidence that matches accuracy (i.e., calibration); pointing out
    ambiguity; declining to answer, or asking a clarifying question, instead
    of inventing a confident answer. These only display the feature, because
    they treat not knowing as a reason not to answer, not as part of working
    on the problem (main text Section~4). Coding agents ask about some badly specified tasks but mostly guess, and when they ask they hand the question to the user. &
    Given a badly specified or unanswerable problem, does the solver point
    this out, consider other framings and hold off, or give a fluent,
    confident answer anyway? This is the ill-posed problem probe (main text
    Section~6). \\

    \rowcolor{zebraCol}
    \rh{4.\ Feedback with the medium} &
    The material being worked on (e.g., a proof, canvas, instrument,
    dataset or codebase) resists the solver and changes the course of the work, so
    that the work responds to the material instead of carrying out a plan
    fixed in advance. &
    A painter responding to what the canvas does; a mathematician led by
    what the proof will and will not permit; adjusting a recipe
    by taste; or rearranging a room to see how it looks. &
    Tool-using agents that revise their plan when the environment
    responds (e.g., error messages, failed tests, retrieved documents). &
    Does what comes back from the environment change the course of the
    work, or does the solver carry on with its original plan? Coding agents pass this test (main text Section~4). A change in understanding of the task is scored under value-laden framing. \\

    \rh{5.\ Value-laden framing} &
    Deciding what counts as a worthwhile problem or a promising move is itself cognitive work and a judgment of value. It depends on the practitioner's developing
    judgment and is not set in advance. &
    Expert judgment of which problems are worth posing and which moves
    are promising; sensing which task on a busy day actually
    matters, or which point in a disagreement is the real one. &
    Largely absent, because the framing comes from the person who writes
    the prompt or sets the goal. AI scientists (systems that automate research) rank hypotheses within a goal
    they are given, which only displays it. Proposed: rewards for reframing
    a problem, not only for solving it (main text Table~1). &
    Does the solver produce or revise its own framing of what matters,
    or optimize a framing handed to it? Look for reframing of the goal,
    not only pursuing a fixed goal or ranking options by fixed criteria. \\

    \rowcolor{zebraCol}
    \rh{6.\ Social and dialogical accountability} &
    Cognition is partly constituted by being answerable to other people,
    reasoning with and against them in dialogue and having one's framing of a problem challenged, and by being answerable for getting things right. &
    \textit{Chavruta} (paired study in which partners argue over a text);
    Socratic dialogue (a claim tested by question and answer); peer review;
    explaining a line of reasoning to a friend who disagrees, or defending
    a plan to colleagues who challenge it. &
    Multi-agent debate, in which agents challenge one another's answers,
    only displays this feature, because answers change mainly through
    agreement, not argument, and the exchanges tend to reduce to a majority
    vote.
    A model that writes both a critique and its reply also only displays it.
    A critic that can change how the problem is framed is proposed (main
    text Section~4). &
    Does a challenge change the framing and the outcome, or do the agents
    agree with one another or with the majority? This is the
    dialogical challenge probe (main text Section~6). \\

    \rh{7.\ Formative dimension} &
    Much of the process cannot be put into words, is built up through
    practice within a community, and changes the practitioner, shaping the
    kind of scientist, artist or professional they become. &
    Becoming the kind of scientist or artist who can do this work; tacit
    skill acquired through sustained practice; growing into a
    parent's judgment, or a driver's feel for the road. &
    Not carried out by deployed systems, whose weights stay fixed between
    releases. Memory files, and self-edited code kept
    by a separate selection process, only display it. Attempted in research
    systems that update their own weights from experience (main text
    Section~4). &
    Does what the solver did change how it approaches the next problem,
    even when its stored record of earlier work is removed? This is the
    sustained-return task of main text Section~6, which for people is repeated over weeks or months. \\

    \bottomrule
  \end{tabularx}
\endgroup
\end{landscape}

\begin{landscape}
\begingroup
  \centering
  \captionof{table}{\textbf{The seven features of process in research.} Each
  feature is given with a short everyday name (in italics). For each
  feature, what it looks like in research, AI use that leaves the feature
  with the researcher, AI use that takes it over, and what record of the
  feature a project can leave for readers, reviewers and journals to
  check (main text Section~5).}
  \label{tab:features-research}
  \fontsize{8.6}{10.2}\selectfont\sffamily
  \renewcommand{\arraystretch}{1.0}
  \setlength{\tabcolsep}{5pt}
  \begin{tabularx}{\linewidth}{@{}>{\raggedright\arraybackslash}p{2.6cm} R K D D@{}}
    \toprule
    \hd{headerDark}{Feature} &
    \hd{humanCol}{In research} &
    \hd{bioCol}{AI use that leaves it with the \mbox{researcher}} &
    \hd{machineCol}{AI use that takes it over} &
    \hd{diagCol}{What a record can show} \\
    \midrule

    \rh{1.\ Generative trial and revision}\newline\textit{Trying and revising} &
    Trying several analyses or framings, and revising them when they fail. In theoretical work, testing an argument against counterexamples and revising it. &
    The tool supplies a hint, a counterexample or the next question, and the researcher makes the next attempt. &
    The tool supplies a finished analysis or argument, so the researcher makes no attempt that could fail. &
    Earlier drafts, version history and abandoned analyses kept in the
    project's files. \\

    \rowcolor{zebraCol}
    \rh{2.\ Temporal extension}\newline\textit{Coming back to it} &
    Returning to the problem over days and months, so that each pass
    reworks the last (e.g., rereading a draft or an analysis after a break). &
    The work is spread across sessions, and in each session the researcher checks whether earlier decisions still make sense. &
    The project is compressed into one session that produces a finished
    output. &
    Dated notes, a decision log, and a preregistration followed by the
    analyses that came after it. \\

    \rh{3.\ Engagement with uncertainty}\newline\textit{Staying with uncertainty} &
    Noticing when a question is badly posed, predicting a result before
    seeing it, and holding off on a conclusion that the data do not yet
    support. &
    The tool points out an ambiguity and leaves it with the researcher
    to resolve. &
    The tool resolves the ambiguity with a confident answer, and the
    researcher accepts it. &
    Predictions written down before the results were seen, stated
    limitations, and questions recorded as still open. \\

    \rowcolor{zebraCol}
    \rh{4.\ Feedback with the medium}\newline\textit{Letting the data change the plan} &
    Reading the codebook (i.e., the document that says what each variable
    and value means), counting the values and plotting the raw data, and
    letting what they show change the analysis plan. &
    The tool helps the researcher look at the data directly, for
    example by producing plots, counts or raw examples. &
    The researcher sees only a summary of the data written by the tool. &
    Exploratory plots and analysis scripts, and each change to the plan
    with the reason for it. \\

    \rh{5.\ Value-laden framing}\newline\textit{Framing the question} &
    Deciding which question is worth asking, and what will count as an
    answer, before the results are in. &
    The researcher poses the question and sets the criteria first, then uses the tool in the analysis. &
    The tool proposes the question and the criteria, and the researcher
    adopts them. &
    Preregistered hypotheses and criteria, and a statement of why the
    question matters. \\

    \rowcolor{zebraCol}
    \rh{6.\ Social and dialogical accountability}\newline\textit{Answering to others} &
    Defending one's choices to collaborators, supervisors and reviewers,
    and changing them when an objection holds. &
    The tool argues against the researcher's position and maintains a sound objection when the researcher disagrees. Colleagues still review the
    work. &
    The tool agrees with the researcher and praises the work (i.e.,
    sycophancy), and it takes the place of discussion with colleagues. &
    Reviewer reports and the authors' responses, and records of the
    discussions in which the work was challenged. \\

    \rh{7.\ Formative dimension}\newline\textit{Becoming a researcher} &
    Building, over years, the judgment to sense that a result is wrong
    before one can say why. &
    The researcher keeps doing the work through which this judgment
    forms (attempting, checking, defending), with the tool in a
    supporting role. &
    The tool does that work, so the occasions on which judgment forms
    are lost and the capacity is not built. &
    Little within a single paper. It shows in the researcher's later work
    and in how they train others. \\

    \bottomrule
  \end{tabularx}
\endgroup
\end{landscape}

\end{document}